\documentclass[conference]{IEEEtran}
\IEEEoverridecommandlockouts

\usepackage{cite}
\usepackage{amsmath,amssymb,amsfonts}
\usepackage{algorithmic}
\usepackage{graphicx}
\usepackage{textcomp}
\usepackage{xcolor}
\usepackage{enumitem}
\usepackage{tabularx}
\usepackage{lipsum}
\usepackage{tabularx}

\def\BibTeX{{\rm B\kern-.05em{\sc i\kern-.025em b}\kern-.08em
    T\kern-.1667em\lower.7ex\hbox{E}\kern-.125emX}}
\begin{document}

\title{Detection of Suicidal Risk on Social Media: A Hybrid Model \\
{\footnotesize \textsuperscript{*}}
}

\author{%
{%
\makebox[\textwidth][c]{%
    \begin{minipage}{0.3\textwidth}%
        \centering
        \textbf{Zaihan Yang}\\
        Suffolk University\\ Boston, USA \\
        zyang13@suffolk.edu
    \end{minipage}%
    \hfill%
    \begin{minipage}{0.3\textwidth}%
        \centering
        \textbf{Ryan Leonard}\\
        Suffolk University\\ Boston, USA \\
        rleonard2@su.suffolk.edu
    \end{minipage}%
    \hfill%
    \begin{minipage}{0.3\textwidth}%
        \centering
        \textbf{Hien Tran}\\
       	Suffolk University\\ Boston, USA \\
        hien.tran@su.suffolk.edu
    \end{minipage}%
}%
}%
\\[1.5em]%
{%
\makebox[\textwidth][c]{%
    \begin{minipage}{0.4\textwidth}%
        \centering
        \textbf{Rory Driscoll}\\
        Suffolk University\\ Boston, USA \\
        rdriscoll@su.suffolk.edu
    \end{minipage}%
    \hfill%
    \begin{minipage}{0.4\textwidth}%
        \centering
        \textbf{Chadbourne Davis}\\
         Suffolk University\\ Boston, USA \\
        chad.davis@su.suffolk.edu
    \end{minipage}%
}%
}%
}
\maketitle

\begin{abstract}
Suicidal thoughts and behaviors are increasingly recognized as a critical societal concern, highlighting the urgent need for effective tools to enable early detection of suicidal risk. In this work, we develop robust machine learning models that leverage Reddit posts to automatically classify them into four distinct levels of suicide risk severity. We frame this as a multi-class classification task and propose a RoBERTa-TF-IDF-PCA Hybrid model, integrating the deep contextual embeddings from Robustly Optimized BERT Approach (RoBERTa),  a state-of-the-art deep learning transformer model, with the statistical term-weighting of TF-IDF, further compressed with PCA, to boost the accuracy and reliability of suicide risk assessment.
To address data imbalance and overfitting, we explore various data resampling techniques and data augmentation strategies to enhance model generalization. Additionally, we compare our model's performance against that of using RoBERTa only, the BERT model and other traditional machine learning classifiers. Experimental results demonstrate that the hybrid model can achieve improved performance, giving a best weighted $F_{1}$ score of 0.7512.
\end{abstract}


\section{Introduction}
Suicidal thoughts and behaviors are increasingly becoming a significant societal concern. As of the latest estimates from World Health Organization, approximately 700,000 to 800,000 people die by suicide globally each year. In the U.S., suicide is the second leading cause of death for individuals aged 10-34 and the fourth leading cause for those aged 35-64. Suicidal thoughts can vary in severity, ranging from explicit and repetitive suicidal feelings, to actively planning suicide or engaging in self-harm behaviors like cutting or burning, and ultimately to making actual attempts through methods such as cutting, jumping, drug overdose, or using firearms.

Various factors influence mental health, leading to serious outcomes such as mental health disorders and suicide of different severity\cite{b1}. However, many of these factors can be mitigated or addressed through effective intervention programs, preventing numerous suicidal tragedies. Efforts have been made from both psychologists and computer scientists to develop effective tools for the early detection and identification of potential suicidal risk to enable timely intervention. For psychologists, they conducted systematic interviews, explored electronic health records to provide solid evidence supporting the findings of potential suicidal factors through statistical analysis\cite{b2, b3, b4}; Computer scientists, in contrast, typically employ machine learning or deep learning techniques using contextual data from social media platforms for suicidal risk detection\cite{b1, b7, b8, b9, b11, b12, b16}.

Social media platforms like Twitter (X), Reddit, Facebook, Tumblr, Weibo and etc, provide a valuable space for individuals to express their thoughts, feelings, and opinions, offering rich data resources for scientists, particularly computer scientists, to develop machine/deep learning models for detecting suicidal ideation and behaviors. As compared to Twitter (X) or other platforms, Reddit provides longer (no word limit), well-structured (with subreddits), and more context-rich discussions, making it an excellent platform for detecting suicidal risk. Its support of anonymity and openness
enables users more likely to share deeply personal struggles without fear of judgment. In spit of these benefits, using social media like Reddit for suicidal risk assessment presents several challenges, including data privacy concerns, the need for large-scale labeled datasets, and the complexity of understanding linguistic nuances across different platforms and cultures. Moreover, the presence of noise, misinformation, and context-dependent meanings in social media posts adds another layer of difficulty in accurately identifying suicidal risk. 
Another significant challenge lies in the varying levels of suicidal ideation—ranging from moderate to severe. Individuals experiencing more severe ideation (e.g., those who have attempted suicide) are often less likely to express their thoughts openly on social media. Our data collection and labeling process supports this observation, resulting in a class imbalance where high-risk cases are underrepresented. This imbalance poses a serious obstacle to building accurate and reliable classification models.

Existing methods for analyzing suicidal risk in social media rely on both traditional machine learning techniques and deep learning approaches, such as transformers. Traditional machine learning techniques, often based on handcrafted features and statistical models, provide interpretability and require relatively fewer computational resources\cite{b13,b25}. However, they struggle with capturing complex linguistic structures and contextual meanings, limiting their effectiveness in nuanced suicide risk prediction. On the other hand, deep learning techniques, particularly transformer-based models like BERT\cite{b29}, RoBERTa\cite{b30}, Gemma and Llama, excel at capturing deep contextual representations of language. These models have demonstrated superior performance in understanding sentiment, detecting suicidal ideation, and processing large-scale text data. Despite their advantages, deep learning models are often criticized for their lack of interpretability, high data dependency, and comparatively high computational costs.

To address these limitations while combining the strengths of both approaches, we propose a hybrid model that integrates word embeddings learned from RoBERTa with TF-IDF (Term Frequency-Inverse Document Frequency). RoBERTa captures deep contextualized representations of words, allowing the model to understand the semantics and nuances of suicidal expressions, while TF-IDF enhances the ability to identify important terms based on their statistical significance across documents. 
The main contribution of our work is highlighted as follows:
\begin{itemize}[leftmargin=*]
\item We proposed a RoBERTa-TF-IDF Hybrid model, integrating the deep contextual embeddings from Robustly Optimized BERT Approach (RoBERTa),  a state-of-the-art deep learning transformer model, with the statistical term-weighting of TF-IDF (Term Frequency-Inverse Document Frequency);
\item To reduce noise and vocabulary dimensionality, thereby enhancing efficiency and model performance, we applied Principal Component Analysis (PCA) to TF-IDF vectors and integrated the reduced representations with word embeddings learned from RoBERTa.
\item We scraped data from Reddit platform, and conducted thorough human labeling, providing sound annotations for posts in terms of their suicidal risk severity;
\item We explored different data resampling as well as data augmentation techniques to deal with data imbalance and overfitting, aiming to obtain a more generalized and robust model;
\item We compared the classification performance using our proposed hybrid model, with that of using RoBERTa only, the BERT model and other traditional machine learning classifiers. Experimental results demonstrate the improved effectiveness of our hybrid model.
\end{itemize}


\section{Literature Review}
In this section, we present a literature review of existing research on suicidal risk detection, focusing on four key aspects: datasets, modeling approaches, feature engineering, and risk severity classification.

Psychologists often rely on patients’ self-reported questionnaires, face-to-face interview transcripts, electronic health records, and other clinical tracking data for suicidal risk analysis\cite{b5, b6, b18}. In contrast, computer scientists typically employ machine learning or deep learning techniques using contextual data from social media platforms. Common platforms used in such studies include Twitter (now X)\cite{b16, b22}, Reddit\cite{b10, b11, b12, b13, b14, b15, b16, b19, b20, b21, b25}, Tumblr\cite{b23}, and the ReachOut Forum\cite{b32}.

The modeling techniques for suicidal risk detection have evolved significantly over time. Early efforts primarily employed traditional machine learning classifiers such as Support Vector Machines (SVM), Logistic Regression, Random Forest, XGBoost and more\cite{b13, b25}. More recent studies have adopted deep learning methods, including Convolutional Neural Networks (CNN), Recurrent Neural Networks (RNN), and Long Short-Term Memory (LSTM) networks\cite{b10, b11, b14, b15, b16, b17, b18, b19, b20, b21}. Currently, transformer-based large language models—such as BERT, RoBERTa, Gemma, GPT, and LlaMA—have become increasingly prevalent due to their superior performance\cite{b12, b19, b22, b26, b27, b28}. While many comparisons have been made across these modeling approaches, few studies, to our knowledge, have attempted to integrate different types of models to harness their complementary strengths. Our work contributes to this relatively unexplored area by exploring such integration.

Feature engineering has played a crucial role in traditional machine learning-based approaches to suicidal risk detection. Common features include word counts, TF-IDF scores, part-of-speech (POS) tags, syntactic and semantic features, n-grams, lexicon-based features, and emotion indicators\cite{b1}. User metadata—such as age, gender, marital status, and occupation—has also been utilized as supportive evidence\cite{b8, b24}. In deep learning, especially with large language models, explicit feature engineering is often less emphasized due to these models’ powerful ability to learn contextual representations. Nonetheless, incorporating feature engineering into deep learning architectures still holds promise for enhancing model performance and interpretability.

Suicidal risk severity has been modeled in various ways. Some studies treat it as a binary classification task—distinguishing between suicidal ideation and non-suicidal content\cite{b13, b16, b18, b19, b20, b21, b22}. Others adopt a multi-class classification framework\cite{b10, b11, b12, b14, b15, b26, b27, b28}, identifying different levels of suicidal risk. For instance, \cite{b14} proposed a four-level categorization: No Risk, Low Risk, Moderate Risk, and Severe Risk. In \cite{b11}, a five-level categorization was used: suicidal indicator, supportive, behavior, and attempt. \cite{b12} later deprecated the "supportive" category from \cite{b11}, as it did not account for users’ comments, and retained the remaining four severity levels.
 

\section{Data Collection and Annotation}
\subsection{Data Set}
Our data set comes from two main resources. One one hand, we obtained the data set provided by researchers from The Hong Kong Polytechnic University\cite{b12}, who scraped posts from 14 subreddits on Reddit, including r/SuicideWatch, r/depression, r/anxiety, r/selfharm, among others. Their provided dataset includes 500 Reddit posts labeled with four different suicide risk levels, along with 1,600 unlabeled Reddit posts. Each post is written in natural English, with punctuations, typos, emojis and some unrecognized non-English characters. 

In addition to the provided data set, we did Reddit scraping by ourselves to collect more data samples. Data collection was performed using Python’s Reddit API wrapper (PRAW), which provides a streamlined interface to Reddit’s data. Posts were scraped from the r/SuicideWatch community by applying multiple sorting strategies ("top", "hot", "controversial", and "new") across diverse temporal intervals (including hourly, daily, weekly, monthly, yearly, and all-time filters). This approach ensured that the data set captured both current trends and historical perspectives, while minimizing redundancy by excluding previously collected posts and handling potential API rate limitations.

Following the initial collection, the data set was filtered down by selecting only posts that featured at least one comment from the original poster, thereby guaranteeing the presence of self-referential context. For each qualifying entry, two distinct representations were created: one that preserved the original post’s title and body, and another that integrated the post with all accompanying responses from the original poster. This strategy facilitates complementary analyses, enabling both a concise and rich, context-aware entry of the post for a larger data set. We finally fetched 899 additional Reddit posts published by 473 unique users, making the entire data set containing 2999 posts in total. All our scraped posts span the period from 2008-12-16 16:03:24 UTC to 2025-01-03 06:01:21 UTC. Table~\ref{tab:dataset} shows a simple statistics of our data set. It is noticable to mention that the word tokens we report here includes all English words, numbers, words connected with underscores, non-English words with Latin letters. Special symbols, punctuation and non-word characters are excluded.


\begin{table}[h]
\centering
\caption{Data Set Statistics}
\label{tab:dataset}
\begin{tabularx}{\linewidth}{X|X|X|X}
  \hline
  \textbf{NO. of posts} & \textbf{NO. of Users} & \textbf{NO. Distinct Word Tokens}  & \textbf{Average NO. of tokens per post} \\ 
  \hline
   2999 & 473 & 13062 & 150.27 \\ 
  \hline
\end{tabularx}
\end{table}


\subsection{Ground-truth human-labeling}
We adopted the labeling scheme and corresponding annotation criteria validated by psychology domain experts, as generally outlined in\cite{b11}. This annotation framework was also employed by researchers at The Hong Kong Polytechnic University on their dataset of 500 Reddit posts\cite{b12}. The criteria are summarized in Table~\ref{tab:risk-level-definition}.

Building on this foundation, we conducted comprehensive annotations on the remaining 2,499 data samples using the same labeling guidelines. The annotation team consisted of five individuals: one faculty member and four undergraduate students, all majoring in Computer Science. The inter-annotator agreement, measured by Fleiss’s Kappa, was 0.5641, indicating moderate agreement.

\begin{table}[h]
\centering
\caption{Definition of Suicidal Risk Levels}
\label{tab:risk-level-definition}
\begin{tabular}{p{0.2\linewidth}p{0.1\linewidth}p{0.5\linewidth}}
  \hline
  \textbf{Category (Risk-Level)} & \textbf{Notation} & \textbf{Definition}  \\ 
  \hline
   Indicator & IN & The post content has no explicit expression concerning suicide \\
   Ideation & ID & The post content has explicit suicidal expression but there is no plan to commit suicide \\
   Behavior & BR & The post content has explicit suicidal expression and a plan to commit suicide or self-harming behaviors \\
   Attempt & AT & The post content has explicit expressions concerning historic suicide \\
  \hline
\end{tabular}
\end{table}


\begin{figure}[htbp]
  \centering
  \caption{Distribution of Suicidal Risk Levels}
  \label{fig:risk-level-distribution}
  \includegraphics[width=0.8\linewidth]{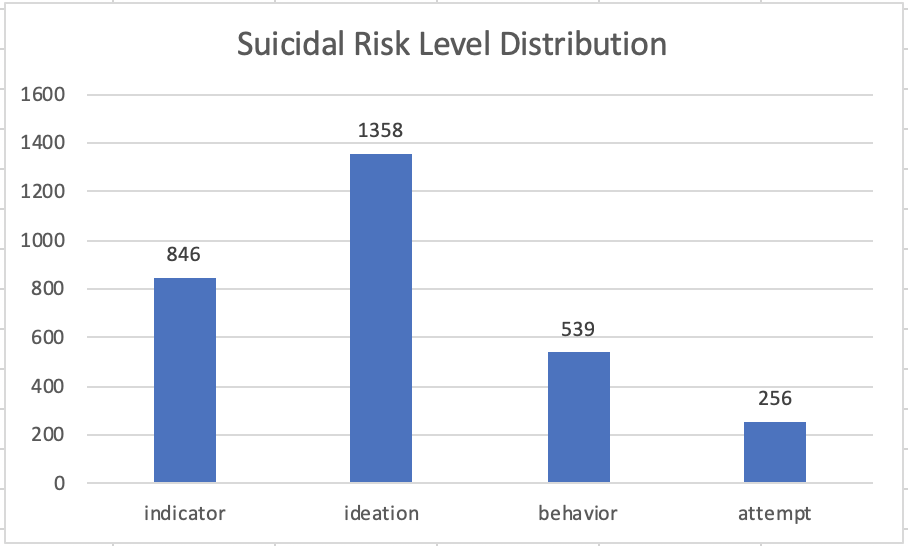}
\end{figure}

Figure~\ref{fig:risk-level-distribution} illustrates the distribution of the four suicidal risk levels in our dataset. As shown, the data is imbalanced, with the ideation category comprising the largest proportion (45.28\%), followed by indicator (28.21\%), behavior (17.97\%), and attempt (8.54\%). This imbalance presents a significant challenge for accurate classification, particularly for categories with limited data, such as attempt. We explored on different data re-sampling as well as augmentation techniques, aiming to address this challenge.





\begin{table}[h]
    \centering
    \caption{Four suicidal risk level Examples}
    \label{tab:risk-level-examples}
    \begin{tabular}{p{8cm}}
        \textbf{\textit{Indicator}}: \textit{People who commit suicide are not weak. Most of the time people get to that place because we've bottled everything up and tried to deal with painful emotions by ourselves then one day we just snap and break down.} \\
        \\
        \textbf{\textit{Ideation}}: \textit{Where can I find out how to die. I don't care if it's slow or painful or both, I just want to definitely die} \\
        \\
        \textbf{\textit{Behavior}}: \textit{"I'm going to kill myself in 6 months if things doesn't get better. I can't be feeling this way forever and I'm so tired of struggling with my mental health. If things doesn't get better then what's the point of being alive."} \\
        \\
        \textbf{\textit{Attempt}}: \textit{I hope I don't make it. I don't want to survive another attempt because all my parents will tell me is how expensive the medical bill is. I just want to die without having to hear them again.} \\
    \end{tabular}
\end{table}
Table~\ref{tab:risk-level-examples} presents examples for each of the four levels of suicidal risk severity. As shown, there are conceptual and linguistic differences among these categories.

\section{Model Design}
Our proposed hybrid model integrates the word-embedding learned from the Robustly Optimized BERT Approach (RoBERTa)\cite{b30}, with the statistical term-weighting of TF-IDF (Term Frequency-Inverse Document Frequency). We also explored different data resampling as well as data augmentation techniques to deal with the problem of imbalanced data and overfitting.

Robustly Optimized BERT Approach (RoBERTa), as an extension to the Bidirectional Encoder Representation from Transformers (BERT) model\cite{b29}, is one of the state-of-the-art deep learning transformer model, which captures deep contextualized word representations, enabling the model to grasp the semantics and nuances of suicidal expressions. However, it may overlook the statistical significance of individual words, which can be crucial for certain classification tasks. To address this, we incorporate TF-IDF-derived features, as TF-IDF excels at identifying key terms based on their statistical significance across documents. Our hybrid model aims at leveraging the strengths of both approaches. 

\subsection{the RoBERTa-TF-IDF-PCA Hybrid Model}
\begin{figure}[h]  
    \centering
    \includegraphics[width=0.46\textwidth]{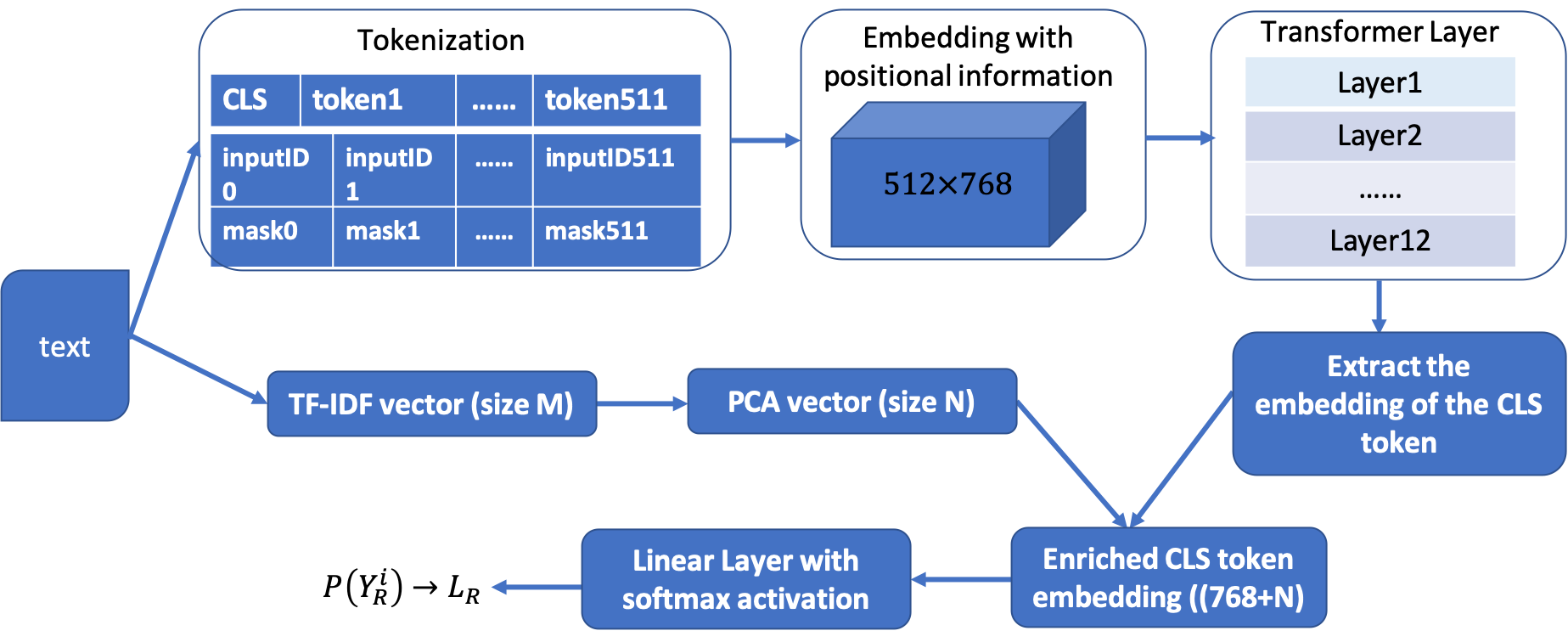}
    \caption{Hybrid Model Framework}
    \label{fig:model-framework}
\end{figure}

The working mechanism of our hybrid model is detailed below and illustrated in Figure~\ref{fig:model-framework}.
\begin{itemize}[leftmargin=*]
\item{\textit{\textbf{Tokenization}}}: RoBERTa applies the byte-level Byte-Pair Encoding (BPE) tokenization scheme to tokenize the raw text of each post into tokens, and convert each token into input IDs with attention masks. Input ID is an unique integer ID based on the pre-trained model's vocabulary and attention mask is a binary tensor (0s and 1s) that indicates which tokens should be attended to by the model. Tokens representing real words are assigned a 1, while padding tokens are assigned a 0. A single \textit{[CLS]} token is added at the beginning of the sequence for the entire post. Unlike BERT, which adds a [SEP] token at the end of each sentence, RoBERTa omits it since Next Sentence Prediction (NSP) is not used during pretraining. If the post is too long (more than 512 tokens as suppored in both BERT-base and RoBERTa-base), additional tokens will get truncated.
\item{\textit{\textbf{Embedding}}}: After tokenization, input IDs are passed through an embedding layer, which maps each token ID to a dense and fixed-length vector representation (embedding) with positional embedding added to provide information about the position of each token in the sequence. In both the RoBERTa and BERT base models, the fixed length is set to be 768. 
\item{\textit{\textbf{Transformer Layers}}}: This embedding (512 × 768 tensor) is then passed through 12 Transformer layers, where it undergoes iterative updates via self-attention and feedforward layers, refining the representation in the context of the entire sequence. The embedding of the \textit{[CLS]} token is finally generated and extracted representing the aggregated information for the entire post.
\item{\textit{\textbf{TF-IDF and PCA Integration}}}: In addition to contextual embeddings from RoBERTa, we incorporate feature vectors derived from TF-IDF representations of the original text. We first represent each tokenized post as a vector of the \textit{M} most significant features based on their TF-IDF scores. To reduce noise and vocabulary dimensionality, we further applied Principal Component Analysis (PCA) to the TF-IDF vectors, selecting the top \textit{N} components that capture the most variance. The resulting \textit{N}-dimensional TF-IDF-PCA vector is then concatenated with the \textit{[CLS]} token embedding obtained from the final transformer layer. This yields a combined feature representation of dimensionality $768+\textit{N}$, where 768 corresponds to the RoBERTa \textit{[CLS]} embedding. In our experiments, we set M be 3000 and N be 300 based on empirical validation.
\item{\textit{\textbf{Output Layer and Loss Function}}}: The enriched post-level embedding of the \textit{[CLS]} token with the size of $768+\textit{N}$ will then be passed through a linear layer followed by a \textit{softmax} activation to predict the probability distribution of different categories of classes (suicide risk levels) $P(Y^i_R)=softmax(linear(\hat{x^i}))$, where the \textit{softmax} function normalizes the output to a [0, 1] range. With $P(Y^i_R) $ computed, categorial cross entropy loss is then adopted for updating of the model parameters, i.e., $L_R=\frac{1}{D}\sum_{i=1}^D\hat{Y^i_R}\cdot\log(P(Y^i_R))$, where \textit{D} is the total number of posts in the training set and $\hat{Y^i_R}$ is the ground-truth label of post \textit{i}. The cross-entropy loss is minimized using the \textit{AdamW} optimizer, which updates the model's parameters through gradient descent.
\end{itemize}

\subsection{Data Resampling}
As indicated in Figure~\ref{fig:risk-level-distribution}, the data is imbalanced for the four categories of different suicidal risk levels. There are much more posts in the "ideation" category which shows general thoughts of suicidal without actual plan or action, than those in the "attempt" category which has taken on real actions. However, it is of great or even more importance to well identify the "attempt" category as it seems more urgent in proper monitoring and intervention. We explored the following three different data re-sampling techniques to investigate whether or not the balanced training data can improve the classification performance on the testing data.
\begin{itemize}[leftmargin=*]
\item{\textit{Over-Sampling (OSam)}}: based upon the category with the largest number of supporting instances (posts), we duplicate the instances of the other categories to achieve a balanced training set.
\item{\textit{Under-Sampling (USam)}}: based upon the category with the smallest number of supporting instances, we randomly sample the instances (posts) of the other categories to achieve a balanced training set.
\item{\textit{Sample Weighted Loss Function (SWL)}}: we use the inverse of the proportion of each category as its weight and construct a weighted loss function. More specifically, the loss function is indicated as $L_R=\frac{1}{D}\sum_{i=1}^D\frac{D}{D_{\hat{Y}^i_R}}\hat{Y^i_R}\cdot\log(P(Y^i_R))$, where $D_{\hat{Y}^i_R}$ is the number of instances in category \textit{i}.
\end{itemize}
 
\subsection{Data Augmentation}
\label{sec:data-augmentation}
Data augmentation in natural language processing involves modifying text while preserving its meaning to improve model robustness. Some data augmentation techniques can increase the sample size as well by synthesizing additional lexically similar samples for the original data corpus. By doing this, the deep learning methods perform training on more or better-presented data samples, hence improving the generalization capacity. The following data augmentation techniques have been explored in our work:
\begin{itemize}[leftmargin=*]
\item{\textit{\textbf{Extending abbreviated words}}}: Reddit posts are often written in natural, conversational English, which includes frequent use of abbreviations. For example, 'kms' stands for 'kill myself,' 'idk' means 'I do not know,' 'lol' refers to 'laugh out loud,' 'ocd' stands for 'obsessive-compulsive disorder,' and 'ptsd' refers to 'post-traumatic stress disorder'. Expanding these abbreviations can enhance the clarity and standardization of post content. To achieve this, we built a dictionary containing 111 abbreviation-expansion pairs and applied it to extend the posts in the training set.
\item{\textit{\textbf{Extending emojis into text}}}: Statistics show that out of our 2,999 posts, 46 contain emojis, a common way for people to express their feelings and moods. To enhance contextual clarity for the learning model, we converted all emojis in the training set into their corresponding English text.
\item{\textit{\textbf{Text summarization}}}: Statistics show that out of a total of 2,999 posts, 142 posts contain more than 512 tokens. In both the RoBERTa-base and BERT-base models, any tokens beyond 512 are truncated, potentially altering or negatively impacting the comprehension of the post's content. To address this limitation, we applied text summarization to extract the main idea of each post while ensuring the token count remains within the 512-token limit.
\item{\textit{\textbf{Using Google Translate}}}: We used Google Translate to translate a post into Spanish and then back into English. This process rephrased the content while preserving its original meaning, introducing synonyms and alternative expressions for words and phrases. As a result, we increased the number of samples in the data set and enriched the contextual representation, both of which contributed to improving the model's generalization. 
\end{itemize}

\section{Experimental Results}

In this section, we present the experimental results using the data set of 2,999 Reddit posts. We compare the classification performance across various data re-sampling techniques, data augmentation methods, and the hybrid model, benchmarking them against the standalone RoBERTa model. Additionally, we evaluate the results obtained using BERT and other traditional machine learning classifiers.
 
\subsection{Experiment Setups}
We applied five-fold cross-validation to obtain stabilized results. The dataset of 2,999 posts was split into five folds without stratification, meaning each fold may have a different distribution of suicidal risk levels. This approach better reflects real-world scenarios.

For each iteration, four out of five folds were combined for training and validation, while the remaining fold was used for testing. Within the training-validation set, we performed an 80:20 stratified split to create the training and validation subsets. The training set was used to train the model, with early stopping applied to prevent overfitting on the validation set. Once trained, the model was evaluated on the testing set.

We assessed classification performance by comparing the predicted suicidal risk levels with the ground truth labels. The results were averaged across the five folds to provide a generalized evaluation of the model's overall performance.


Due to the imbalanced distribution of suicidal risk levels in our dataset (as shown in Figure~\ref{fig:risk-level-distribution}), we used weighted precision, weighted recall, and weighted $F_{1}$ scores as evaluation metrics to account for class imbalance.

\subsection{Experiment Results}
We made several groups of comparisons in terms of classification results between different models, data resampling techniques and data augmentations.

\subsubsection{Data Resampling}
During five-fold cross-validation, we applied oversampling or undersampling to both the training and validation sets to achieve data balance, ensuring that each suicidal risk level had an equal number of supporting evidence (posts); we also evaluated on using the reciprocal of the proportion of each risk level as its weight to construct a weighted loss function. The data distribution for the test set remained unchanged across all experiments. Table~\ref{tab:results-data-resample} presents the performance results. "original" means no data-resampling applied, keeping the original data distribution in the training and validation set.

\begin{table}[h]
    \centering
    \caption{Comparision of Different Data Resampling Techniques}
    \label{tab:results-data-resample}
    \begin{tabular}{c|c|c|c}
        \hline
         & \textbf{weighted precision} & \textbf{weighted recall} & \textbf{weighted $F_1$ score} \\ 
        \hline
        Original & 0.7523 & \textbf{0.7496} & \textbf{0.7499} \\ 
        OSam & 0.7485 & 0.7439 & 0.7421 \\ 
        USam & 0.7120 & 0.7009 & 0.7022 \\ 
        SWL & \textbf{0.7524} & 0.7486 & 0.7480 \\ 
        \hline
    \end{tabular}
\end{table}

As indicated, Oversampling (OSam), Sample Weighted Loss function (SWL), and the Original method achieve competitive results, with the Original method obtaining the highest weighted $F_1$ score of 0.7499, followed by SWL and OSam. One possible explanation for this is that the dataset was split into five folds without stratification, leading to varying distributions of suicidal risk levels across folds.

Oversampling balanced the different risk levels within the training set, but this may have caused a mismatch with the imbalanced distribution in the test set, potentially affecting generalization. Similarly, SWL assigns weights inversely proportional to the number of posts per risk level; however, since the test set follows a different distribution, the learned weights may not generalize well.

Undersampling produced the worst results, likely because it reduced the number of training samples, thereby limiting the model’s ability to learn effectively and degrading overall performance.
  

\subsubsection{Data Augmentation}
In this group of comparison, we incorporated all four data augmentation techniques (with DA) we introduced in Section~\ref{sec:data-augmentation} into RoBERTa learning process, and compared the classification results against the model without data augmentation (w/o DA). Oversampling was adopted as the data resampling technique in this comparison experiment. Table~\ref{tab:results-data-augmentation} shows the results.


\begin{table}[h]
    \centering
    \caption{With vs. Without Data Augmentation}
    \label{tab:results-data-augmentation}
    \begin{tabular}{p{1.5cm}|p{1.5cm}|p{1.5cm}|p{1.5cm}}
        \hline
         & \textbf{weighted precision} & \textbf{weighted recall} & \textbf{weighted $F_1$ score} \\ 
        \hline
        w/o DA & 0.7485 & 0.7439 & 0.7421 \\ 
        \hline
        DA & 0.7015 & 0.6589 & 0.6648 \\ 
        \hline
    \end{tabular}
\end{table}
As indicated, incorporating data augmentation techniques did not improve classification performance, as the weighted precision, recall, and $F_1$ scores all decreased. Several factors may contribute to this decline: Expanding abbreviations and emojis can lead to unnatural sentence structures that are misaligned with RoBERTa’s learned embeddings. Since abbreviations and emojis often carry contextual or sentiment cues, expanding them may dilute or distort their meaning; Text summarization may slightly alter the original intent of a post, potentially affecting classification accuracy; Google Translate can introduce inaccuracies or unnatural synonyms, making sentences less representative of the original text. Further investigation is needed to analyze the impact of each augmentation technique individually.

\subsubsection{Hybrid Model}
We compared and report the results comparing the performance of our hybrid model with that of using word embedding from RoBERTa only, and with the BERT model. Table~\ref{tab:results-hybrid-model} shows the results.


\begin{table}[h]
    \centering
    \caption{Performance of RoBERTa-TF-IDF-PCA: the Hybrid Model}
    \label{tab:results-hybrid-model}
    \begin{tabular}{p{2cm}|p{1.5cm}|p{1.5cm}|p{1.5cm}}
        \hline
         & \textbf{weighted precision} & \textbf{weighted recall} & \textbf{weighted $F_1$ score} \\ 
        \hline
        RoBERTa-TF-IDF-PCA & \textbf{0.7557} & \textbf{0.7532} & \textbf{0.7512} \\ 
        \hline
        RoBERTa & 0.7523 & 0.7496 & 0.7499 \\ 
        \hline
        BERT & 0.7045 & 0.6996 & 0.6949 \\
        \hline
    \end{tabular}
\end{table}

For both the RoBERTa-TF-IDF-PCA hybrid model and the standalone RoBERTa model, we did not perform any re-sampling to balance the dataset. Instead, we retained the original class distribution, as it yielded the best performance, as shown in Table~\ref{tab:results-data-resample}. No data augmentation techniques were applied either, since they negatively impacted performance under the current experimental setup.

Regarding hyperparameters, we set the batch size to 3, the learning rate to 1e-5, and the weight decay to 0.01 to apply L2 regularizationn in these experiments. These values were chosen based on empirical testing. Additionally, we employed the DistilBERT model for this experiment as one baseline comparison model.

The results highlight the effectiveness of the hybrid model, which achieved the highest weighted scores across all three evaluation metrics—precision, recall, and $F_1$ score. Notably, the highest weighted $F_1$ score achieved was 0.7512.


\begin{table}[h]
    \centering
    \caption{Classification results for four severity levels}
    \label{tab:results-four-levels}
    \begin{tabular}{p{1.8cm}|p{1cm}|p{1cm}|p{1cm}|p{1cm}}
        \hline
        	\textbf{Model} & \textbf{Indicator-$F_{1}$} & \textbf{Ideation-$F_{1}$} & \textbf{Behavior-$F_{1}$} & \textbf{Attempt-$F_{1}$} \\
        \hline
        RoBERTa-TF-IDF-PCA & 0.7541  & 0.7961 & 0.6767 & 0.6527 \\
        \hline
        RoBERTa & 0.7681 & 0.7875 & 0.6721 & 0.6437 \\
        \hline
    \end{tabular}
\end{table}

Table~\ref{tab:results-four-levels} presents the average $F_1$ score for each of the four severity levels across five folds using the hybrid RoBERTa-TF-IDF-PCA model. As shown, for three out of four severities, the hybrid model outperform the model with RoBERTa embedding only. Moreover, performance aligns with the number of supporting posts for each level: the \textit{ideation} category achieves the highest performance, followed by \textit{indicator}, \textit{behavior}, and \textit{attempt}. This highlight the impact of class imbalance, and indicate that  more dedicated modeling strategies or features may be required to capture their unique characteristics.

\begin{figure}[h]  
    \centering
    \includegraphics[width=0.40\textwidth]{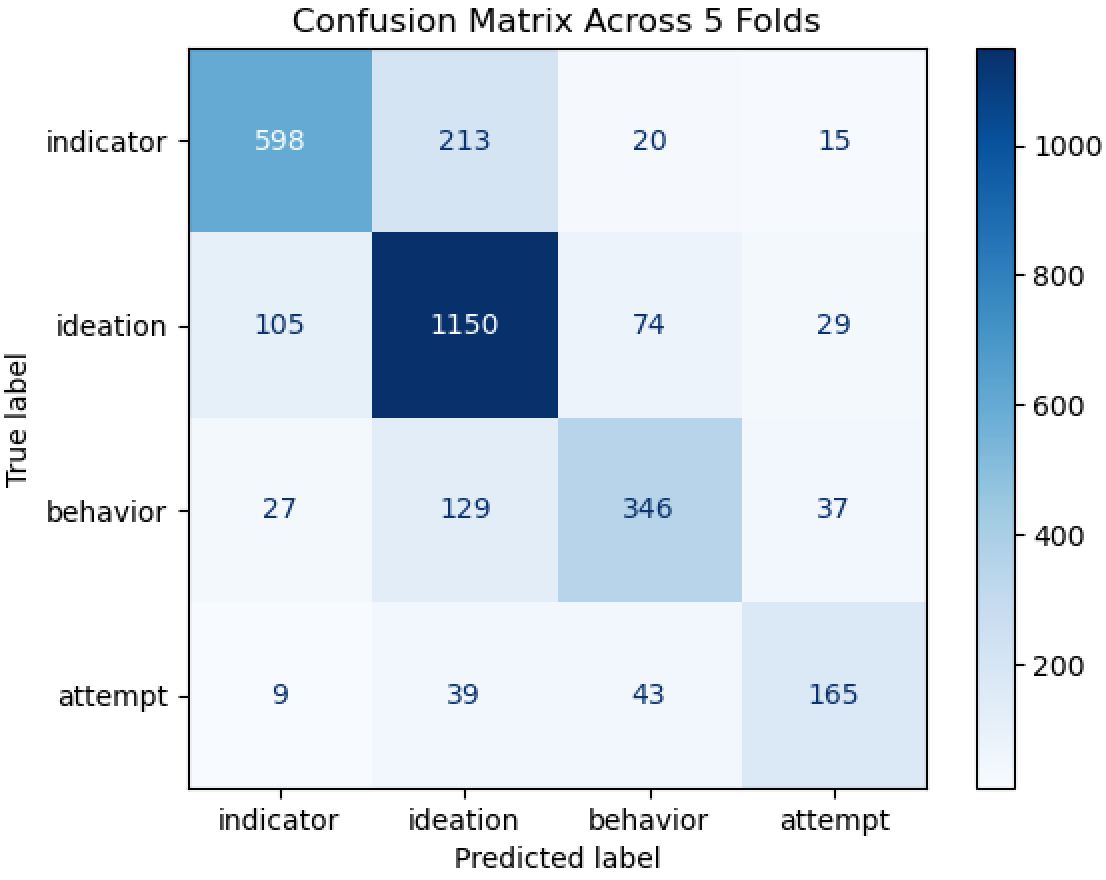}  
    \caption{Confusion Matrix for Classifier}
    \label{fig:conf_matrix}
\end{figure}


Figure~\ref{fig:conf_matrix} shows the Confusion Matrix across five folds using the hybrid RoBERTa-TF-IDF-PCA model with the original categorical distribution. As indicated, the model frequently confuses \textit{indicator} with \textit{ideation} (213 instances) and vice versa (105 instances). \textit{behavior} is also commonly misclassified as \textit{ideation} (129 instances). The greatest confusion for \textit{attempt} occurs with \textit{behavior} (43 instances) and \textit{ideation} (39 instances). These patterns suggest that the model struggles to distinguish between ideation, behavior, and attempt, likely due to conceptual or linguistic similarities among these categories. To further improve classification performance, it may be necessary to incorporate additional training samples and identify more discriminative features.

\subsubsection{Comparison with Traditional Machine Learning Models}
Before the rise of deep learning models—particularly transformer-based architectures like BERT and RoBERTa—traditional machine learning classifiers demonstrated strong performance across a wide range of classification tasks. In this study, we evaluated their effectiveness on the specific task of suicidal risk detection and compared their performance against the RoBERTa model. We selected four traditional classifiers: Support Vector Machine (SVM), Logistic Regression, Naive Bayes, and Random Forest, each applied to two types of feature vector representations:
\begin{itemize}[leftmargin=*]
\item{\textit{TF-IDF}:} in which posts are first tokenized, and represented as a vector of the 3,000 most significant features based on their TF-IDF scores.
\item{\textit{Word2Vec}:} in which posts are first tokenized and then represented as vectors of real numbers in a continuous vector space using Word2Vec\cite{b31}, one of the popular nature language processing techniques using shallow neural networks to learn word associations from a large corpus of text. In such a vector space, words with similar meanings are located close to each other. We set the vector size be 1000 here in this experiment.
\end{itemize}

Using TF-IDF feature vectors, we applied and compared various data pre-processing techniques—such as stopword removal, stemming, and lemmatization—to assess their impact on classification performance. The results, shown in Table~\ref{tab:results-other-classifiers}, were obtained using five-fold cross-validation and evaluated with the weighted $F_1$ score as the performance metric.

\begin{table*}[ht]
    \centering
    \caption{Comparison with traditional classifiers}
    \label{tab:results-other-classifiers}
    \begin{tabular}{|c|c|c|c|c|}
        \hline
        	& \textbf{SVM} & \textbf{Logistic Regression} & \textbf{Naive Bayes} & \textbf{Random Forest} \\
        \hline
        TF-IDF & 0.5849  & 0.5658 & 0.3073 & 0.4717 \\
        \hline
        TF-IDF + remove stopwords & 0.5727 & 0.5637 & 0.3227 & 0.5033 \\
        \hline
        TF-IDF + stemming & 0.5880 & 0.5725 & 0.3061 & 0.4805 \\
        \hline
        TF-IDF + lemmatization & 0.5825 & 0.5725 & 0.3082 & 0.4711 \\
        \hline
        Word2Vec & 0.4347 & 0.4901 & 0.3387* & 0.4831 \\
        \hline
    \end{tabular}
\end{table*}

As indicated in Table~\ref{tab:results-other-classifiers}, several observations can be made: 1) SVM consistently outperformed the other traditional classifiers across all data pre-processing techniques, followed by Logistic Regression, Random Forest, and Naive Bayes. It’s worth noting that we used Gaussian Naive Bayes with Word2Vec features, and Multinomial Naive Bayes with the TF-IDF variants. 2) TF-IDF-based feature representations yielded better performance than Word2Vec across the classifiers. 3) The impact of pre-processing techniques varied by classifier. For instance, stemming improved performance for SVM and Logistic Regression, while stopword removal boosted results for Naive Bayes and Random Forest. 4) Compared with the results in Table~\ref{tab:results-hybrid-model}, RoBERTa outperformed all traditional classifiers, further highlighting the effectiveness of transformer-based models for this task.

\section{Summary and Future work}

Detecting varying levels of suicidal ideation risk is crucial for providing timely and effective interventions, which can potentially reduce suicide rates. In this work, we approach the task as a multi-class classification problem and propose a hybrid RoBERTa–TF-IDF-PCA model. Our approach integrates word embeddings learned by RoBERTa with TF-IDF feature vectors within the RoBERTa transformer framework. To address issues of data imbalance and overfitting, we experimented with various data resampling strategies and data augmentation techniques. Additionally, we evaluated classification performance using traditional machine learning classifiers with different feature representations and compared them against the RoBERTa-based models. Experimental results demonstrate that our proposed hybrid model outperforms the other baselines, achieving the highest weighted precision, recall, and $F_1$ scores, with a top weighted $F_1$ score of 0.7512.


Future work can be explored in the following directions:
\begin{itemize}[leftmargin=*]
\item{Integrating hybrid techniques}: Investigate more effective methods for combining the strengths of deep learning and traditional machine learning approaches—such as ensemble techniques like bagging or boosting to improve model generalization, especially for poorly classified samples.
\item{Feature representation combinations}: Explore the classification performance of alternative or additional combinations of feature vector representations beyond RoBERTa embeddings and TF-IDF vectors.
\item{Alternative deep learning models}: Evaluate the performance of other deep learning architectures, including LSTM, as well as emerging transformer-based models such as Gemma and LlaMA.
\item{Incorporating richer textual context}: Assess the impact of integrating supplementary textual content, such as follow-up comments or related posts from other subreddits, into the classification process.
\item{Temporal user-level analysis}: Analyze the progression of suicidal ideation risk levels over time based on posts from the same user. This could offer deeper insight into key triggers and turning points in users' mental health status, enabling more timely and personalized interventions.
\end{itemize}

\end{document}